\documentclass[letterpaper, 10 pt, conference]{ieeeconf}  

\usepackage{graphicx}
\usepackage{amsmath}

\IEEEoverridecommandlockouts                              

\usepackage{color}
\usepackage{subfig}

\usepackage{amssymb}
\usepackage{pifont}
\newcommand{\cmark}{\ding{51}}%
\newcommand{\xmark}{\ding{55}}%

\usepackage{bm}
\DeclareMathOperator*{\argmax}{arg\,max}

\title{\LARGE \bf
Interpretable and Flexible Target-Conditioned Neural Planners For Autonomous Vehicles
}

\author{Haolan Liu$^{1}$ Jishen Zhao$^{2}$ Liangjun Zhang$^{3}$
\thanks{$^{1}$Haolan Liu is with the University of California San Diego, {\tt\small hal022@ucsd.edu}. Work done as an intern at Baidu Research}
\thanks{$^{2}$Jishen Zhao is with the University of California San Diego, 
        {\tt\small jzhao@ucsd.edu}}
\thanks{$^{3}$Liangjun Zhang is with Baidu Research, Sunnyvale, CA 94089, USA {\tt\small liangjunzhang@baidu.com}}%
}

\begin{document}

\maketitle
\thispagestyle{empty}
\pagestyle{empty}

\begin{abstract}

Learning-based approaches to autonomous vehicle planners 
have the potential to scale to many complicated real-world driving scenarios by leveraging huge amounts of driver demonstrations. However, prior work only learns to estimate a single planning trajectory, while there may be multiple acceptable plans in real-world scenarios. To solve the problem, we propose an interpretable neural planner to regress a heatmap, which effectively represents multiple potential goals in the bird's-eye view of an autonomous vehicle. The planner employs an adaptive Gaussian kernel and relaxed hourglass loss to better capture the uncertainty of planning problems. We also use a negative Gaussian kernel to add supervision to the heatmap regression, enabling the model to learn collision avoidance effectively. Our systematic evaluation on the Lyft Open Dataset across a diverse range of real-world driving scenarios shows that our model achieves a safer and more flexible driving performance than prior works.


\end{abstract}

\section{INTRODUCTION}


The past decade has witnessed a continuous proliferation of autonomous vehicle (AV) research and industry practice \cite{ChaufferNet-19,Yurtsever-survey-2020,DBLP:journals/corr/abs-2103-01882}. 
Among all the AV subtasks including perception and motion forecasting, the planning task is especially challenging due to the complicated real-world driving scenarios and dynamic interaction with other traffic agents.
Traditional approaches typically formulate the planning task as a cost optimization problem, in a predefined parameterized trajectory space (e.g., cubic spirals~\cite{lattice}). However, such approaches require tremendous efforts in fine-tuning the cost functions and other hyperparameters, which is not scalable and cost-effective~\cite{DBLP:journals/corr/abs-1910-04586}.
The learning-based approaches leverage real-world expert demonstration to learn the ideal driving behavior. 
Such data-driven approaches can easily scale to a diverse range of driving scenarios~\cite{ChaufferNet-19, Zeng2019EndToEndIN,DBLP:journals/corr/abs-2103-01882}.

\begin{figure}[htp]

\subfloat[The multilanes scenario.]{%
  \includegraphics[width=0.5\textwidth]{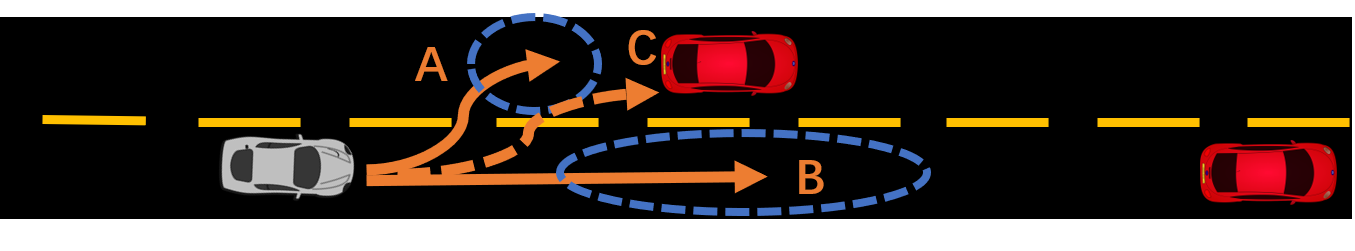}
  \label{fig:scenario1}
}

\subfloat[The nudging scenario.]{%
  \includegraphics[width=0.5\textwidth]{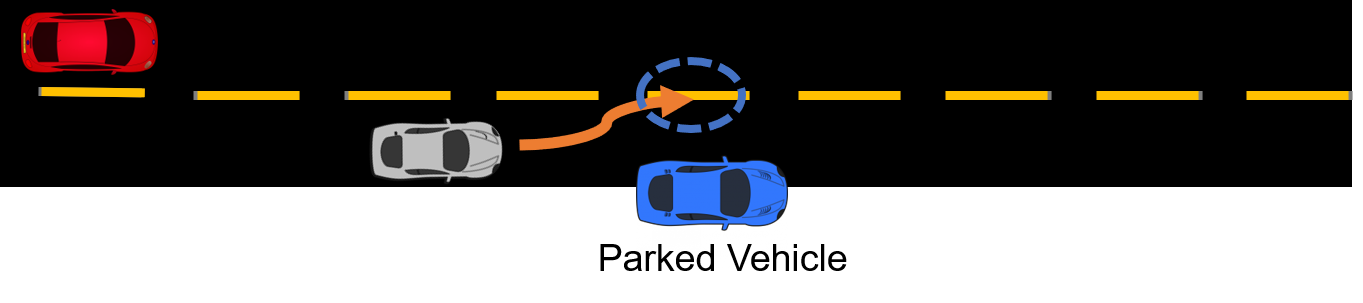}
  \label{fig:scenario2}
}

\caption{(a) A multilane scenario: the AV has multiple possible trajectories, but the dataset may only include a single ground-truth trajectory for each data sample. (b) A nudging scenario: a blue vehicle is parked near the curb. AV needs to learn to nudge a little bit and safely pass the blue vehicle.}
\label{fig:scenario}
\end{figure} 





However, pure imitation-based approaches have their limitations due to distribution shifts, which result in compounding errors due to the sequential nature of driving decisions~\cite{ChaufferNet-19}. In particular, the existing imitation learning approaches are mostly restricted to regress an optimal trajectory based on the current driving scenario~\cite{ChaufferNet-19, DBLP:journals/corr/abs-2103-01882}. 
However, the optimal trajectory inherently involves uncertainties. 
Figure~\ref{fig:scenario1} shows an AV (the grey vehicle, whose routing information is to turn left at the next intersection) that has two possible driving plans, driving into the left turn lane (goal A) or passing the red vehicle and trying to change lane (goal B). Yet we can only observe one of them for each sample in the training dataset.
Such unimodal regression may learn to predict an average of two modes~\cite{Cui-2019}, leading to risky plans (dotted line C may hit the rear end of the red vehicle).

Moreover, even when the only acceptable action is switching to the left lane -- there are still ambiguities in the labels.
As the groundtruth trajectory still can be affected by individual driving styles.
For example, a driver may drive faster and keep a longer following distance than others. We depict the variance as the blue dotted ellipse. 
Such uncertainty in the ground-truth labels needs to be captured to improve learning performance. 
To make things worse, the variance of different plans can also be different. Figure~\ref{fig:scenario2} shows a nudging scenario with a badly parked blue vehicle, where an AV needs to perform a flexible nudging maneuver, instead of getting stuck. The variance of this maneuver is smaller compared with Figure~\ref{fig:scenario1}, as the AV also cannot borrow the other lanes too much, which may hamper the normal traffic. To support such flexible maneuvers, the learning-based planner needs to adaptively capture the variance of different plans.



To address those limitations, we propose to learn a probability distribution of acceptable planned trajectory, rather than focusing on the optimal trajectory. 
Inspired by works in target-driven motion forecasting, we make the observation that the value/cost/optimality of a trajectory can be mostly captured by the goal it wants to reach.
Specifically, we regress an interpretable heatmap representation, indicating which goals are preferable in the map.
Our model takes the input of mid-level representation from perception and outputs future waypoints of ego vehicles. 
In summary, we conclude our major contributions in the following:

\begin{itemize}
    \item Our model regresses an interpretable heatmap that predicts the value of different goal positions on the map. Our auxiliary task also helps the AV predict the trajectories of other vehicles, which explicitly improves the collision avoidance ability of AV planners.
    \item We provide relaxed hourglass regression loss to force the model to focus on the important region of the heatmap. We also propose an adaptive Gaussian kernel to capture the uncertainty of the planning problem and the inherent labeling ambiguity. 
    \item We demonstrate our model to achieve better safety and flexibility compared with prior works, by evaluating them with a diverse range of realworld driving scenarios from the large-scale Lyft dataset. 
\end{itemize}

\section{BACKGROUND AND RELATED WORK}


\subsection{Imitation Learning}
The imitation learning approach for learning driving policies using large amounts of driver demonstrations suffers from the distribution shift problem~\cite{Pomerleau-1989-15721}, which can result in dangerous driving behaviors over time. 
The ChauffeurNet~\cite{ChaufferNet-19} augments the driving data by perturbing the position or velocity, which makes the AV robust to the error induced by distribution shift. It also demonstrates its capability to drive in a closed-loop control environment. 
To mitigate the mode averaging problem, InfoGAIL combines imitation learning with the generative model to capture multi-modal behavior of the expert demonstration~\cite{DBLP:journals/corr/LiSE17}.

Meanwhile, it is desirable for learning-based systems to provide interpretable results that help us understand the decisions they make. The neural motion planner~\cite{Zeng2019EndToEndIN} attempts to do so by learning an interpretable cost volume supervised by expert demonstrations and other constraints, such as collision avoidance and traffic rules.



\subsection{Reinforcement Learning (RL) and Inverse RL}
Reinforcement learning (RL) specifies a performance metric (reward), while the agent learns the optimal behavior by interacting with the environment. However, designing the reward function (reward shaping) is difficult for the complicated real-world driving tasks~\cite{Wang-RL-Lane-Change-18}. To this end, inverse reinforcement learning (IRL) attempts to identify the reward function from expert demonstrations~\cite{huang2021driving}.
However, all those approaches suffer from the sim-to-real gaps that may render the learned policy performing poorly in the real-world settings.
Besides, the learned representation is not interpretable, making it hard to apply in safety-critical AV applications.

\subsection{Target-conditioned Prediction}
Motion forecasting task predicts the plausible future trajectory sets based on likelihood fitting or generative models~\cite{Cui-2019,DBLP:journals/corr/LeeCVCTC17}. Recent motion forecasting approaches adopt a target-conditioned motion model to handle the intrinsic multimodality of motion forecasting~\cite{DenseTNT-21}. 
Mixture Density Network (MDN) is also used to model the multimodal distribution of the motion forecasting tasks~\cite{Cui-2019}. However, MDN training is brittle and unstable as it heavily relies on good initialization. It also easily suffers from mode collapse~\cite{MDN-hard}.

\begin{figure}[htp]
\centering

\includegraphics[width=0.28\textwidth]{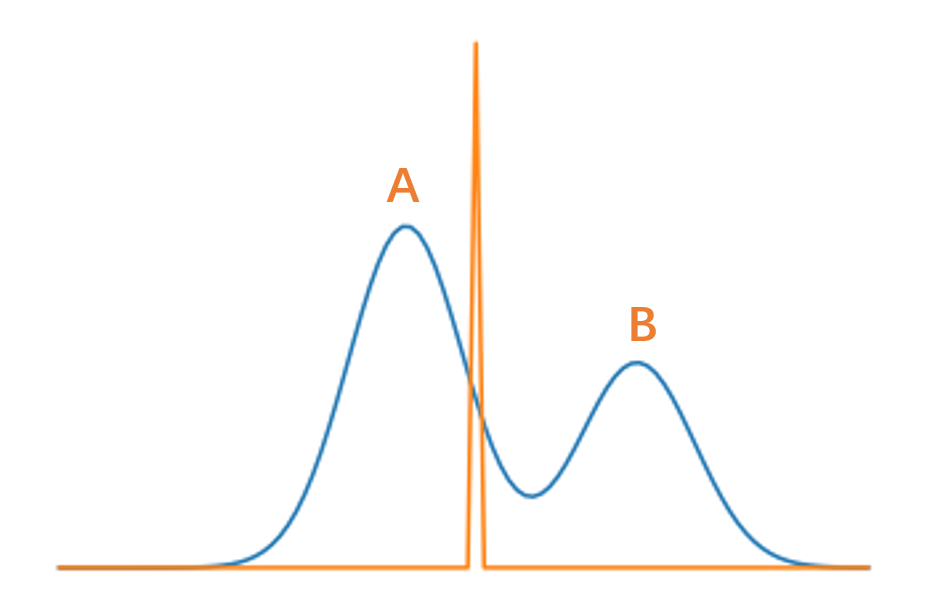}

\caption{ The probability distribution of the target is often multimodal: the distribution of this scenario has two modes A and B, which corresponds to the plan A and B in Figure~\ref{fig:scenario1}. Prior work such as ChauffeurNet learns a single optimal trajectory, based on L1/L2 loss. The regressed trajectory suffers from mode averaging, leading to suboptimal policy.}
\label{fig:scenario_distribution}
\end{figure}

\section{Methods}

\begin{figure*}[t]
    \center{
    \includegraphics[width=0.8\textwidth]{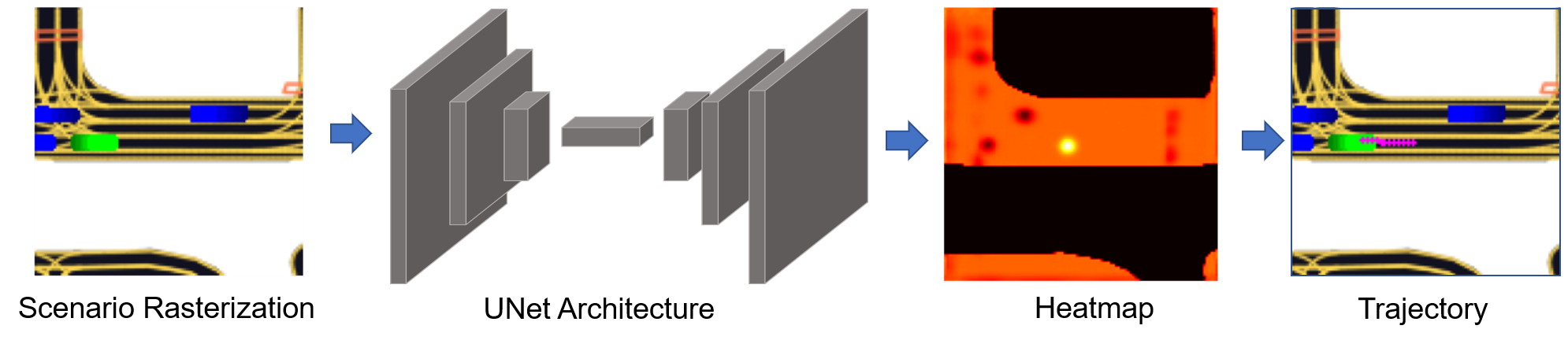}}
    \caption{Our planning model takes the input of multi-channel rasterization image from bird's-eye view. The first stage network is a UNet architecture that outputs a heatmap of the same size with the rasterization image, indicating the value of each position. Given the heatmap, we use argmax operation to find the coordinate with the highest value in the heatmap. The second stage network will take the goal and scenario embedding, and output the planning trajectory (waypoints).}
    \label{fig:overview}
\end{figure*}


The purpose of motion planning is to plan a future trajectory for the ego vehicle (AV), for $T$ timesteps ($\bf{s}$=($x_t$,$y_t$,$\theta_t$) for t in 1,..., $T$, $x$, $y$, $\theta$ indicates the 2D position and the orientation of the AV). The model is given the observation
of the past $N$ frames, the observation ($\bf{x}$) includes the map information, the dynamic traffic agents including vehicles and pedestrians, and the history position of the ego vehicle. Therefore, we want to capture the probability distribution $p(\bf{s}|\bf{x})$, for the planning problem. Such a problem is similar to the motion forecasting task. Previous work points out that uncertainty can be decomposed into intent uncertainty and control uncertainty~\cite{DBLP:journals/corr/abs-2008-08294}.
As the driving intention is determined by the AV itself, prior work assumes that the probability distribution of optimal trajectory is unimodal.
For example, ChauffeurNet and its variation use L1 or L2 imitation loss, with the assumption of unimodal Laplacian or Gaussian noises~\cite{ChaufferNet-19, DBLP:journals/corr/abs-2103-01882}.


In our work, we propose to learn the value of multiple acceptable trajectories, instead of regressing a single trajectory.
Inspired by prior work, we work on the planning goals of the AV, rather than regressing a single high-dimensional trajectory. 
We assume that the value of a planning goal is mostly captured by its goals. To formulate our framework, the value of a goal $\bf{g}$ is learned by function $\bf{V_\tau}$=$f(\bf{\tau,x})$. To reduce computation, we discretize the driving scenario and compute all the $\tau={(i,j)}$ out of the rasterization image space.  Then we select the optimal goal via $\tau=\argmax{(V_{ij})}$. With the optimal goal, we will regress a trajectory $T=\nu(\tau)$ for the AV to reach that goal.

\begin{figure}[ht]
  \subfloat[Agent Boxes]{
	\begin{minipage}[c][1.0\width]{
	   0.15\textwidth}
	   \centering
	   \includegraphics[width=1\textwidth]{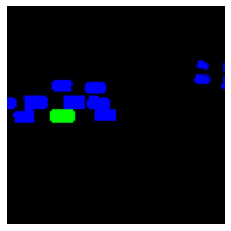}
	\end{minipage}}
 \subfloat[Map Information]{
	\begin{minipage}[c][1.0\width]{
	   0.15\textwidth}
	   \centering
	   \includegraphics[width=1\textwidth]{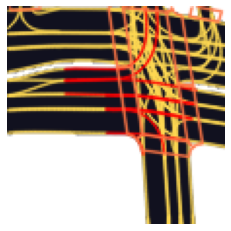}
	\end{minipage}}
 \hfill	
  \subfloat[Navigation Mask]{
	\begin{minipage}[c][1.0\width]{
	   0.15\textwidth}
	   \centering
	   \includegraphics[width=1\textwidth]{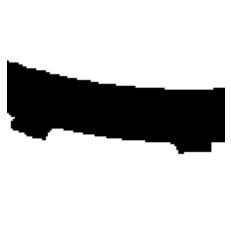}
	\end{minipage}}
 
 \hfill 	
  
\caption{The scenario rasterization in bird's-eye view (BEV): (a) the multi-channel bounding boxes of traffic agents, including several history frames and the current frame to describe the ego vehicle (AV) and the past agents' trajectory. We visualize the history trajectory with sequences of fading brightness. (b) the map information, typically including the lanes, traffic lights and crosswalk. (c) the navigation mask shows the ego's routing information. }
\label{fig:input}
\end{figure}

Observation $\bf{x}$ is multi-channel rasterized bird’s-eye view (BEV) images as input, which are rendered based on the groundtruth data in this paper. As shown in Figure~\ref{fig:input}, the rasterized image includes the last $n$ history frames and the current map information, including traffic lights, road topology, and navigation direction.
The BEV intermediate representation precisely captures the environment information including location and scale of objects and is widely used in previous works~\cite{https://doi.org/10.48550/arxiv.2203.17270, bev-modnet, bev-seg, bev-voxel}.

\subsection{Model Architecture} 

Our planner firstly predicts a heatmap with the same size of BEV images, each pixel's value ranges from 0 to 1, presenting the optimality of the position. During training, we will compute the loss between the predicted heatmap and the groundtruth heatmap, constructed with 2D Gaussian kernels.
During inference, the AV will pick the position with the maximum heatmap value as the short-term goal. 
Conditioned on that, the next stage network will predict a sequence of future waypoint ($x$,$y$,$yaw$) and continuously control the AV. 

Our first stage network adopts the ResNet18 and UNet architecture, which is originally designed for semantic segmentation tasks~\cite{unet-15, resnet}. We select the UNet architecture due to its capability to capture spatial relationships, which is important in the motion planning task. 
The second stage is implemented by a two-layer MLP network, that takes the environment embedding and the predicted goals and output a trajectory to reach that goal.

\subsection{Heatmap Regression}
Compared with directly regressing the numerical coordinates of the points of interest, the heatmap regression is proved to have better spatial generalization~\cite{DBLP:journals/corr/abs-1801-07372}.
Therefore, heatmap regression is widely used to regress coordinates for tasks such as keypoints detection and bounding box detection~\cite{DBLP:journals/corr/abs-2012-15175,bulat2017far}. 
The groundtruth heatmaps are constructed with 2D Gaussian kernels, centered on the labeled points.

\begin{equation}
\begin{split}
G(x,y)=\frac{1}{2\pi\sigma^{2}}e^{-\frac{(x-x_0)^2+(y-y_0)^2}{2\pi \sigma ^{2}}}   \\
\textit{s.t.} \| x-x_0 \|_1 \leq 3 \sigma ,  \| y-y_0 \|_1 \leq 3 \sigma
\end{split}
\end{equation}

The ($x_0$, $y_0$) is the groundtruth short-term goals. 
The heatmap models the optimality of the goals instead of the probability, therefore we remove the normalization coefficient $\frac{1}{2\pi\sigma^{2}}$.
$\sigma$ indicates the uncertainty of the goal. The value range of every pixel is 0-1.

The heatmap regression approach in previous work~\cite{DBLP:journals/corr/abs-2012-15175,bulat2017far} has multiple Gaussian kernels as the groundtruth, which handles the multimodal distribution naturally. 
However, as we only observe a single trajectory in the dataset, the groundtruth heatmap only has one mode. To encourage the model to learn multiple acceptable goals, we introduce the negative Gaussian kernel and relaxed hourglass loss.

\paragraph{Negative Gaussian Kernel}

\begin{figure}[ht]
  \subfloat[The bird's-eye view of original scenario]{
	\begin{minipage}[c][1.0\width]{
	   0.15\textwidth}
	   \centering
	   \includegraphics[width=1\textwidth]{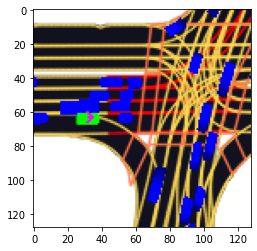}
            \label{fig:scenario_without_bb}
	\end{minipage}}
 \subfloat[The object mask]{
	\begin{minipage}[c][1.0\width]{
	   0.15\textwidth}
	   \centering
	   \includegraphics[width=1\textwidth]{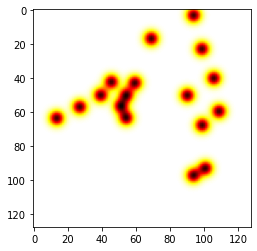}
        \label{fig:obj_mask}
	\end{minipage}}
  \subfloat[The road mask]{
	\begin{minipage}[c][1.0\width]{
	   0.15\textwidth}
	   \centering
	   \includegraphics[width=1\textwidth]{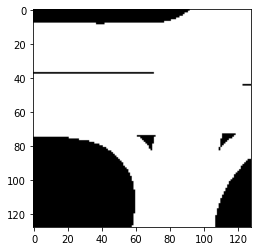}
    \label{fig:road_mask}
	\end{minipage}}

\caption{We add the road mask and object mask into the groundtruth heatmap, to supervise the neural planners to learn to drive on road and avoid collision.}
\label{fig:mask}
\end{figure}

To keep safe, the goals that may collide with other traffic agents should have a lower value. Inspired by that, we initialize the groundtruth heatmap to be 0.5 everywhere. Then we add the positive Gaussian kernel centered on the goal and the negative Gaussian kernel centered on the object position after the planning horizon.
Both kernels are normalized by 0.5, which makes the groundtruth position on the heatmap the highest value 1, and other pixels still have a value range 0-1.
We also mask the off-road region on the rasterized image as 0. We visualize those negative Gaussian kernel and road mask in Figure~\ref{fig:obj_mask} and Figure~\ref{fig:road_mask}.

\begin{figure}[ht]
 \subfloat[$\sigma=5$]{
	\begin{minipage}[c][1.0\width]{
	   0.15\textwidth}
	   \centering
	   \includegraphics[width=1\textwidth]{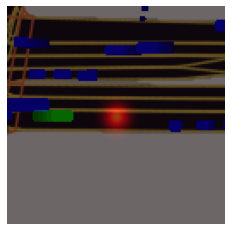}
            \label{fig:sigma5}
	\end{minipage}}
 \hfill
  \subfloat[$\sigma=2$]{
	\begin{minipage}[c][1.0\width]{
	   0.15\textwidth}
	   \centering
	   \includegraphics[width=1\textwidth]{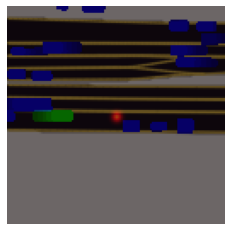}
        \label{fig:sigma2}
	\end{minipage}}
 \subfloat[]{
 \begin{minipage}[c][1.0\width]{
	   0.15\textwidth}
	   \centering
	   \includegraphics[width=1\textwidth]{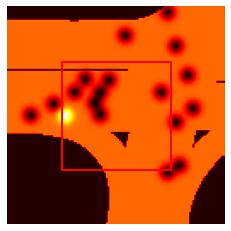}
            \label{fig:heatmap_gt}
	\end{minipage}}
 
 \caption{(a), (b): groundtruth with different $\sigma$ values; (c) The groundtruth heatmap for the scenario in Fig. ~\ref{fig:mask} constructed by object mask, road mask and the groundtruth goal. We only calculate the L2 loss for pixels inside the red line patch.}

\end{figure}

By adding the objects' future position into the groundtruth heatmap, our model is learning to predict their future position. Due to the limitation of the receptive field, we cannot correctly predict the object position that is not initially in the receptive field. Therefore, we only calculate the loss for the center patch of the heatmap, as shown in Figure~\ref{fig:heatmap_gt}. This design also reduces the number of the pixel needed to compute the loss, which may alleviate the imbalanced distribution problem in heatmap regression~\cite{DBLP:journals/corr/abs-2012-15175}.

In our experiment, we also observe that our AV sometimes will change lanes to avoid high-speed following vehicles, which is not desirable. Therefore, we filter the objects that are driving behind the AV and also in the same lane when we compute the object negative Gaussian kernel.

\paragraph{Adaptive Gaussian Kernel}

In previous work, different goals are set up with the same standard deviation~\cite{DBLP:journals/corr/abs-2012-15175}. Such assumption doesn't hold true for the motion planning domain. Figure~\ref{fig:scenario1} shows that the variance of different goals in the driving scenarios exhibit different variance: the variance of goal A is small compared with goal B, because of the nearby red vehicle in the neighbor lane. 
Capturing the variance is critical to the safety, as some scenarios (e.g., ~\ref{fig:scenario2} requires very small variance in the planning goals, which means the maneuvers should be very accurate. 
Therefore, we adaptively set the standard deviation of the Gaussian kernel based on different scenarios. Specifially, we sample different standard deviation $\sigma$ from 1 to 5, and use the largest $\sigma$ that doesn't incur collision. Figure~\ref{fig:sigma5} and Figure~\ref{fig:sigma2} shows the process, which helps our model adaptively capture the uncertainty in different scenarios.

\paragraph{Relaxed Hourglass Loss}
The heatmap regression task typically uses the Mean Squared Loss (MSE)~\cite{DBLP:journals/corr/abs-2012-15175}.
However, with an increasing number of pixels, prior work and our experiment show that the negative pixels can overwhelm the entire loss function, hampering the neural network from learning useful hints. 
As mentioned, our groundtruth heatmap labels the drivable but suboptimal region with the value 0.5. However, this is overly strict as the drivable region is relatively large compared with the groundtruth kernel and object kernels. Therefore, we downweight the loss for the drivable region with following loss function:

\begin{align}
L_{regression}= \| \bm{W} \| \cdot \| \bm{\hat{Y}-Y} \|^2_2 \\
W_{ij}=  \left\{\begin{matrix}
        0.6 & if  \hat{y_{ij}}  = 0.5 \\
        1 & otherwise
    \end{matrix}\right.
\end{align}

The $\bm{Y}$ and $\bm{\hat{Y}}$ refer to the predicted and groundtruth heatmap value, respectively.
$\| \bm{W} \|$ stands for the downweighting matrix that adjusts the MSE Loss. $\|\|_2$ stands for L2 norm.
We name it hourglass loss as the loss function puts more weight on the two ends (0 and 1) and downweights the intermediate value.
Another benefits of downweighting is to alleviate the penalty for the multimodal distribution shown in Figure~\ref{fig:scenario_distribution}, which can improve the performance of our learned motion planners.

\subsection{Implementatin Details}
\label{subsection:impl_detail}



\paragraph{Dataset \& Scene Renderers}
We train our model with a large-scale real-world driving dataset, Lyft dataset~\cite{DBLP:journals/corr/abs-2006-14480}.
The dataset was collected over multiple cities, with a diverse range of road conditions. It consists of around 170000 scenarios, each about 25 seconds long.
We are using the built-in rasterization module in Lyft dataset to render the rasterized image of the scenarios. L5kit renders our AV (i.e., the ego vehicle) at the (0.25, 0.5) of the image space, and the initial orientation always aligns with x axis of the image space.
The input image is 128x128 and the rendering resolution is 0.5 meter per pixel. Therefore, AV can sense the spatial region spanning 48 meters in front and 16 meters in back, 32 meters to the left and right. The planning frequency is 10 frames per second, and the planner output a trajectory with 2 seconds duration. Our models are trained using the Adam optimizer with an initial learning rate of 0.0001.

\paragraph{Data Augmentation}

We examine the goal distribution of Lyft dataset and find that Lyft driving dataset is highly skewed with driving straight cases. Such unbalanced distribution may hamper the generalizability~\cite{imbalanced}. It can also lead AV fail to change lanes when necessary, hurting the flexibility of the neural planners. 
We divide the training dataset scenarios into three parts, namely ``going straight", ``turning left" and ``turning right". For example, for the ``turning left" scenario, the maximum shift of vehicle orientation in the planning horizon is over a threshold (we use 0.4 in radians) in the left direction. Our result shows the ratio of three categories is about 48:1:1.
To balance their distribution, we adjust the sampling frequency of the scene data, by downweighting the frequency of ``going straight" cases.
Such reweighting cases can balance the training distribution, forcing our model to learn flexible behaviors such as lane changing.
In addition, we add a small uniform-distributed noise (-$\frac{\pi}{6}$,$\frac{\pi}{6}$) to the rendering orientation, making the 
AV's orientation not necessarily points to the x-axis of the image coordinate.

\begin{figure}[ht]
  \subfloat[The original trajectory]{
	\begin{minipage}[c][1.0\width]{
	   0.15\textwidth}
	   \centering
	   \includegraphics[width=1\textwidth]{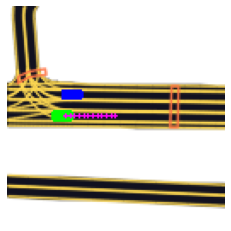}
              \label{fig:pre-perturb}
	\end{minipage}}
 \hfill
 \subfloat[The perturbed trajectory]{
	\begin{minipage}[c][1.0\width]{
	   0.15\textwidth}
	   \centering
	   \includegraphics[width=1\textwidth]{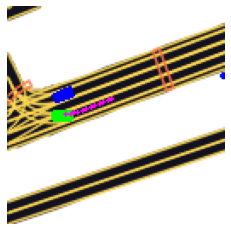}
            \label{fig:perturb}
	\end{minipage}}
   \subfloat[The heatmap of the out-of-distribution scenario]{
	\begin{minipage}[c][1.0\width]{
	   0.15\textwidth}
	   \centering
	   \includegraphics[width=1\textwidth]{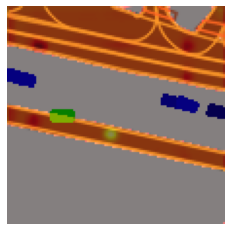}
        \label{fig:perturbed_heatmap}
	\end{minipage}}
 
\caption{(a) shows the original scenario. (b) shows the scenario after perturbation: the initial position and orientation is perturbed a little, and the rest of the trajectory is interpolated. Such perturbation can teach the planner to drive on lanes.
(c) the heatmap visualization shows that our model can learn to get back on lanes even with large perturbation noise.
}
\label{fig:perturbation}
\end{figure}

\paragraph{Perturbation}
Imitation learning in sequential decision problems usually suffers from error accumulation, as it leads to the states that are very poorly distributed in the training dataset. To cope with this, we adopt the same data augmentation method as in ChauffeurNet, by mutating the position and orientation of some frames in the trajectory. 
Figure~\ref{fig:perturbation} lists an example that the orientation and initial position of the vehicle is perturbed, then the rest of the waypoints are interpolated based on the fixed end point. Such synthesized cases can expose learned motion planner to some states that don't exist in the expert demonstration (e.g., driving offroad and collision), which is proved to be effective in addressing the error accumulation issues~\cite{ChaufferNet-19}.
We use the perturbation probability of 0.1.

\section{RESULTS}

\begin{table*}[]
\centering
\begin{tabular}{|c|c|c|c|c|c|}
\hline
 &
  Description &
  Loss function &
  \begin{tabular}[c]{@{}c@{}}Interpretable\\ Representation\end{tabular} &
  \begin{tabular}[c]{@{}c@{}}Collision Cases\\ /Rear end cases\end{tabular} &
  Total Fail Cases \\ \hline
M0  & \begin{tabular}[c]{@{}c@{}}The ChauffeurNet \\ implementation from L5kit.\end{tabular} & L2 loss                & \xmark                 & 11/9  & 13 \\ \hline
M1  & Neural motion planner                                                                             & max margin loss        & \cmark & 27/18 & 21 \\ \hline
M2a & Vanilla heatmap regression                                                                        & L2 loss                & \cmark           & 59/22 & 32 \\ \hline
M2b & \begin{tabular}[c]{@{}c@{}}Heatmap regression with\\ adaptive Gaussian kernel\end{tabular}        & L2 loss                & \cmark           & 52/23 & 29 \\ \hline
M2c & \begin{tabular}[c]{@{}c@{}}Heatmap regression with \\ road and object mask\end{tabular}           & relaxed hourglass loss & \cmark           & 18/12 & 24 \\ \hline
M2d & \begin{tabular}[c]{@{}c@{}}Heatmap regression with \\ all our contributions\end{tabular}          & relaxed hourglass loss & \cmark           & 8/5   & 4  \\ \hline
\end{tabular}
\caption{Comparison of prior work and different model configuration.}
\label{tab:data}
\end{table*}

\subsection{Evaluation Methodology}

\paragraph{Driving Scenarios}
Motion planners need to be both efficient and safe across a diverse range of real-world driving scenarios. 
We first preprocess the data from Lyft testing dataset~\cite{DBLP:journals/corr/abs-2006-14480} and select representative scenarios based on their categories, which are:
\begin{itemize}
    \item \textbf{Lane Following}: the ego vehicle (AV) is driving in straight or curved roads without lane changing, the main challenge is to keep a reasonable speed, and also not be too closed to preceding vehicles
    \item \textbf{Lane Changing}: including merging lanes, lane changing behaviors. The ego vehicle needs to correctly overtake or yield.
     \item \textbf{Intersection}: The ego vehicle will perform a right turn or left turn to the correct lane, and also correctly behave based on the traffic signal.
      \item \textbf{Flexibility Testing}: We also curated some scenarios with badly parked vehicles (nudging scenario in Figure~\ref{fig:scenario} (b)) or preceding vehicles with very slow speeds. AV should learn to overtake those vehicles safely.

\end{itemize}

We collect 1000 testing scenarios, each with 15 seconds, from the dataset. We also balance their frequency in the four categories for testing diversity and coverage. 
We perform closed-loop simulation with our AV planners by replaying the dataset, including the recorded trajectory of non-AV vehicles and actuate the waypoints generated by our model. We also collect necessary metrics such as collision rates.

\paragraph{Evaluation Metrics}
To evaluate our planner, we use similar metrics in the previous work: (1) the collision rate: we collect the number of collision cases in the simulation. (2) the pass/fail rate. We also select 50 scenarios with specific requirements, such as nudging, lane merging, and yielding. We observe the behaviors of the AV and mark the scenario as ``Fail" if it fails to perform ideal behaviors, such as getting stuck in the traffic or deviating from the predefined route. 

\paragraph{Comparison And Ablation Analysis}
We compare our neural planners with two prior works, ChauffeurNet~\cite{ChaufferNet-19} and NMP~\cite{Zeng2019EndToEndIN}, in Table~\ref{tab:data}. Note that the NMP planner takes the input of raw sensor data, we retrain a NMP planner that takes the input of rasterized images for a fair comparison.
We also perform an ablation study with different configurations (M2a-d) to demonstrate the effectiveness of our design.

\begin{table}[]
\centering
\begin{tabular}{|c|c|c|c|c|}
\hline
             & \begin{tabular}[c]{@{}c@{}}Lane \\ Following\\ (8 cases\\  in total)\end{tabular} & \begin{tabular}[c]{@{}c@{}}Lane \\ Changing\\ (13 cases \\ in total)\end{tabular} & \begin{tabular}[c]{@{}c@{}}Intersection\\ (11 cases \\ in total)\end{tabular} & \begin{tabular}[c]{@{}c@{}}Flexibility \\ Testing\\ (18 cases\\ in total)\end{tabular} \\ \hline
ChauffeurNet & 8                                                                                 & 13                                                                                & 11                                                                             & 5                                                                                      \\ \hline
NMP          & 7                                                                                 & 11                                                                                 & 8                                                                             & 3                                                                                      \\ \hline
Our model    & 8                                                                                 & 13                                                                                & 10                                                                            & 15                                                                                     \\ \hline
\end{tabular}
\caption{The pass cases in the 4 categories for different models.}
\label{tab:passfail}
\end{table}

\begin{table}[]
\centering
\begin{tabular}{|c|c|c|c|}
\hline
          & ChauffeurNet & NMP  & Our work \\ \hline
Pass Rate & 63\%         & 48\% & 94\%     \\ \hline
\end{tabular}
\caption{The pass rates in out-of-distribution scenarios for different models.}
\label{tab:ood}
\end{table}

\subsection{Safety Analysis}
We test different model configuration in Table~\ref{tab:data}. We can conclude that our model can achieve better safety promises than prior work, as it causes fewer collision rates (comparing M2d with M1 and M0). We also find that the object mask (comparing M2c with M2a) helps the AV learn to predict the trajectory of other vehicles, which significantly reduce the collision cases.

We also list the pass/fail cases in Table~\ref{tab:passfail} in four categories of our 50 scenario testing challenges. We can see that our work is much better than NMP in all 4 categories. Compared with ChauffeurNet, our model also achieves comparable performance in the first three normal traffic scenarios (go straight, lane changing, and intersection). Our model exhibits much better performance in the flexibility testing category.
Our visualization in Figure~\ref{fig:small_variance} and Figure~\ref{fig:large_variance} shows different variance in the goal heatmap, Figure~\ref{fig:small_variance} is smaller due to the existence of other traffic agents. It shows that our model can capture the variance of driving goals, which explains its ability to perform flexible maneuvers.

As we are only replaying the log in our closed-loop simulators, the simulated agent in the rear of ego vehicle will not react to the slowdown behavior. We also list the number of rear-end collision cases in the table, in which the AV is not responsible for the collisions.

\subsection{Generalization Evaluation}
One major problem of the imitation learning approach is the distribution shift. In this section, we evaluate the generalization ability in two kinds of scenarios: (1) perturbing the initial position and velocity of the ego vehicle with larger offsets and variance compared with the data augmentation mentioned in Section~\ref{subsection:impl_detail}. 
(2) perturbing the trajectory of the other traffic agents. We also filter the unrealistic scenarios such as deliberate collision. Our requirements are that the AV should avoid collisions and quickly recover to the lanes.
Table~\ref{tab:ood} shows the pass rates of our model and prior works, which proves our work can better generalize to out-of-distribution scenarios.
We also visualize a perturbed scenario in Figure~\ref{fig:perturbed_heatmap}. Although such drastic deviations from the lanes are not likely in the training dataset, our model still learns to get back on lanes.

\subsection{Visualization}
\begin{figure}[]
\centering
  \subfloat[The original scenario]{
	\begin{minipage}[c][1.0\width]{
	   0.14\textwidth}
	   \centering
	   \includegraphics[width=1\textwidth]{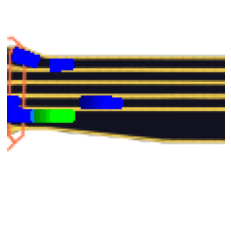}
	\end{minipage}}
 \subfloat[The predicted heatmap]{
	\begin{minipage}[c][1.0\width]{
	   0.14\textwidth}
	   \centering
	   \includegraphics[width=1\textwidth]{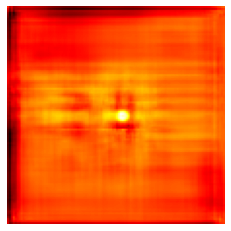}
	\end{minipage}}

    \caption{(a) one multi-lane driving scenario. (b) The predicted heatmap of vanilla heatmap regression for (a), with a single Gaussian kernel centered on the groundtruth position. }
         \label{fig:viz_vanilla}

 \end{figure}

\begin{figure}[]
\centering
  \subfloat[]{
	\begin{minipage}[c][1.0\width]{
	   0.14\textwidth}
	   \centering
	   \includegraphics[width=1\textwidth]{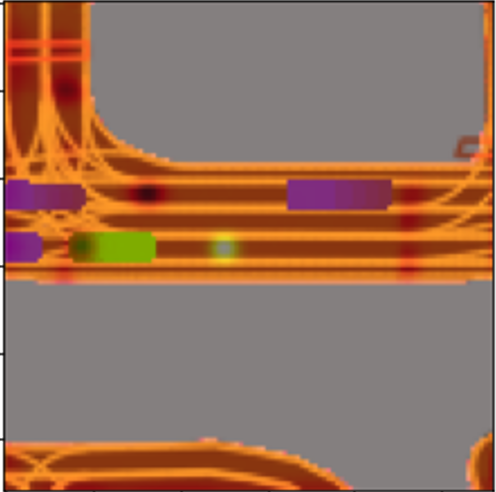}
    \label{fig:heatmap1}
	\end{minipage}}
 \subfloat[]{
	\begin{minipage}[c][1.0\width]{
	   0.14\textwidth}
	   \centering
	   \includegraphics[width=1\textwidth]{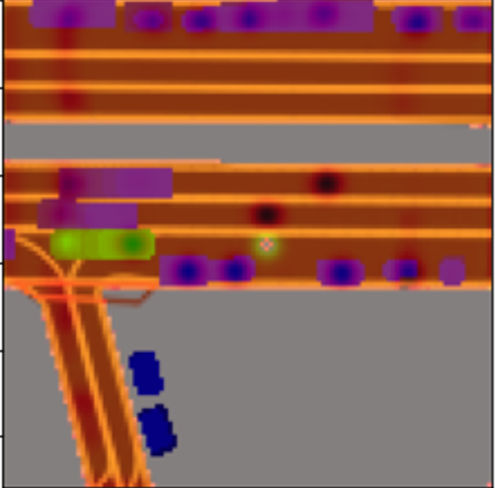}
     \label{fig:small_variance}
	\end{minipage}}
  \subfloat[]{
	\begin{minipage}[c][1.0\width]{
	   0.14\textwidth}
	   \centering
	   \includegraphics[width=1\textwidth]{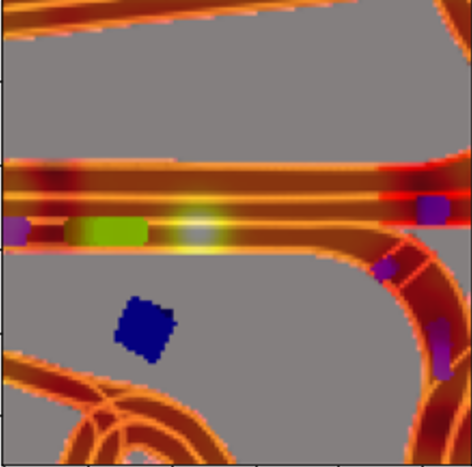}
        \label{fig:large_variance}
	\end{minipage}}
    
    \caption{We overlay the predicted heatmap on different testing scenarios. The bright region indicates acceptable goals, while the dark region indicates the region that may involve collision.}
    \label{fig:viz}
 \end{figure}

The predicted heatmap of our model indicates the value of different planning goals. We first visualize the heatmap of the vanilla heatmap regression (M2a) in Figure~\ref{fig:viz_vanilla}. We can see that the highlight region is reasonable and safe for a planning goal. 
In comparison, Figure~\ref{fig:viz} shows the heatmap of our model. We can see that the model can accurately capture the drivable region and optimal goals (highlight region) in the scenarios. The dark region also shows the motion forecasting result of other vehicles. Figure~\ref{fig:viz} (a) even shows multiple dark region for the prediction of the rightmost vehicle, which can be explained by that the preceding vehicle may perform a lane change. Another example is that Figure~\ref{fig:viz} (c) shows that the heatmap value is very low in the intersection with red light signal, which demonstrates the learned rules of stopping at a red light signal.

\subsection{Performance Analysis}
We extensively evaluated the runtime performance of our method on a machine with NVIDIA Tesla P100 GPU. The model inference time is 11 ms, and the BEV rendering time is 23.94 ms.
As a comparison, ChauffeurNet takes 160 ms on the same GPU device~\cite{ChaufferNet-19}.

\section{Conclusion}
This paper proposes a learning-based planner that takes the input of bird’s-eye view (BEV) rasterized images, and outputs an interpretable heatmap that indicates the value of different goals on the map. Our framework explicitly captures the uncertainty in the planning problem by adaptive standard deviation of the 2D Gaussian kernel. We also use the negative Gaussian kernel for traffic agents to build the ability to avoid collisions. We propose the relaxed hourglass loss to encourage capturing multiple modes of acceptable goals, while still keeping the ability to select optimal planning goals and keep safe. Our evaluation on a large-scale real-world dataset shows our planner is safer and more flexible compared with prior imitation learning planners.






\clearpage


\bibliographystyle{abbrv}
\bibliography{references}

\end{document}